\documentclass[letterpaper, 10 pt, conference]{ieeeconf}  % Comment this line out if you need a4paper

\IEEEoverridecommandlockouts                              % This command is only needed if 
\usepackage{algorithm}
\usepackage{algpseudocode}
\usepackage{graphicx}

\usepackage[dvipsnames]{xcolor}
\usepackage{graphics}
\usepackage{amsmath}

\usepackage{amssymb}
\usepackage{epsfig} % seems important for fig 1 at least
\usepackage{booktabs}
\usepackage{float}
\usepackage{url}
\usepackage{gensymb}
\usepackage{xspace}
\def\OURS{MetaEMG\xspace}
\newcount\Comments % 0 suppresses notes to selves in text
\usepackage{color}
\definecolor{darkgreen}{rgb}{0,0.5,0}
\definecolor{purple}{rgb}{1,0,1}
\newcommand{\kibitz}[2]{\ifnum\Comments=1\textcolor{#1}{#2}\fi}
\title{\LARGE \bf
Meta-Learning for Fast Adaptation in Intent Inferral \\ on a Robotic Hand Orthosis for Stroke

}
\author{Pedro Leandro La Rotta$^{*, 1}$, Jingxi Xu$^{*, 2}$, Ava Chen$^1$, Lauren Winterbottom$^3$, Wenxi Chen$^1$, \\Dawn Nilsen$^{3,4}$, Joel Stein$^{3,4}$, and Matei Ciocarlie$^{2,4}$ % <-this % stops a space
\thanks{$^*$These authors have contributed equally to this work.}%
\thanks{$^{1}$Department of Mechanical Engineering,
        Columbia University in the City of New York, NY, USA
        {\tt\small \{pll2127, ava.chen, wc2746, matei.ciocarlie\}@columbia.edu}}%
\thanks{$^{2}$Department of Computer Science, Columbia University in the City of New York, NY, USA
        {\tt\small jxu@cs.columbia.edu}}%
\thanks{$^{3}$Department of Rehabilitation and Regenerative Medicine, Columbia University Irving Medical Center, New York, NY 10032, USA
{\tt\small \{lbw2136, dmn12, js1165\}@cumc.columbia.edu}}%
\thanks{$^4$ Co-Principal Investigators}
}

\begin{document}

\maketitle
\thispagestyle{empty}
\pagestyle{empty}

%%%%%%%%%%%%%%%%%%%%%%%%%%%%%%%%%%%%%%%%%%%%%%%%%%%%%%%%%%%%%%%%%%%%%%%%%%%%%%%%
\begin{abstract}

We propose \OURS, a meta-learning approach for fast adaptation in intent inferral on a robotic hand orthosis for stroke. One key challenge in machine learning for assistive and rehabilitative robotics with disabled-bodied subjects is the difficulty of collecting labeled training data. Muscle tone and spasticity often vary significantly among stroke subjects, and hand function can even change across different use sessions of the device for the same subject. We investigate the use of meta-learning to mitigate the burden of data collection needed to adapt high-capacity neural networks to a new session or subject. Our experiments on real clinical data collected from five stroke subjects show that \OURS can improve the intent inferral accuracy with a small session- or subject-specific dataset and very few fine-tuning epochs. To the best of our knowledge, we are the first to formulate intent inferral on stroke subjects as a meta-learning problem and demonstrate fast adaptation to a new session or subject for controlling a robotic hand orthosis with EMG signals. 

\end{abstract}

%% body 
\section{INTRODUCTION}

Intuitive and robust interfaces for user-driven control of wearable robots have the potential to help enhance therapy inside and outside of the clinical setting, across a wide spectrum of health conditions. However, the development of such interfaces has proven difficult, even with recent advances in computing and machine learning. 

In upper-limb devices, user-driven interfaces often employ a model that predicts user intent from surface electromyography (EMG) data. Such a model needs to not only capture the relationship between signal and intent across individuals, but it also needs to be robust to some amount of variation from factors like user fatigue and sensor migration. 
In the stroke survivor demographic, hand spasticity and muscle tone can also change drastically across different sessions (single use of the device between donning and doffing) on different days (commonly known to the community as intersession concept drift), adding an additional challenge for model development in the stroke rehabilitation space.

While utilizing complex models to have a better chance of generalization across the large variation in data is a promising research direction, data scarcity in the stroke survivor demographic is a major limiting factor. Both collecting and annotating data are burdensome for therapists and stroke survivors, and very few open-source datasets currently exist. This restricts our ability to address the challenges discussed above by simply learning high-capacity intent inferral models trained on large amounts of data. While more data always helps, we believe that the field also stands to benefit from the investigation of alternative, complementary approaches that enable us to to train larger models with potentially less data. 

\begin{figure}
    \centering
    \includegraphics[width=1\linewidth]{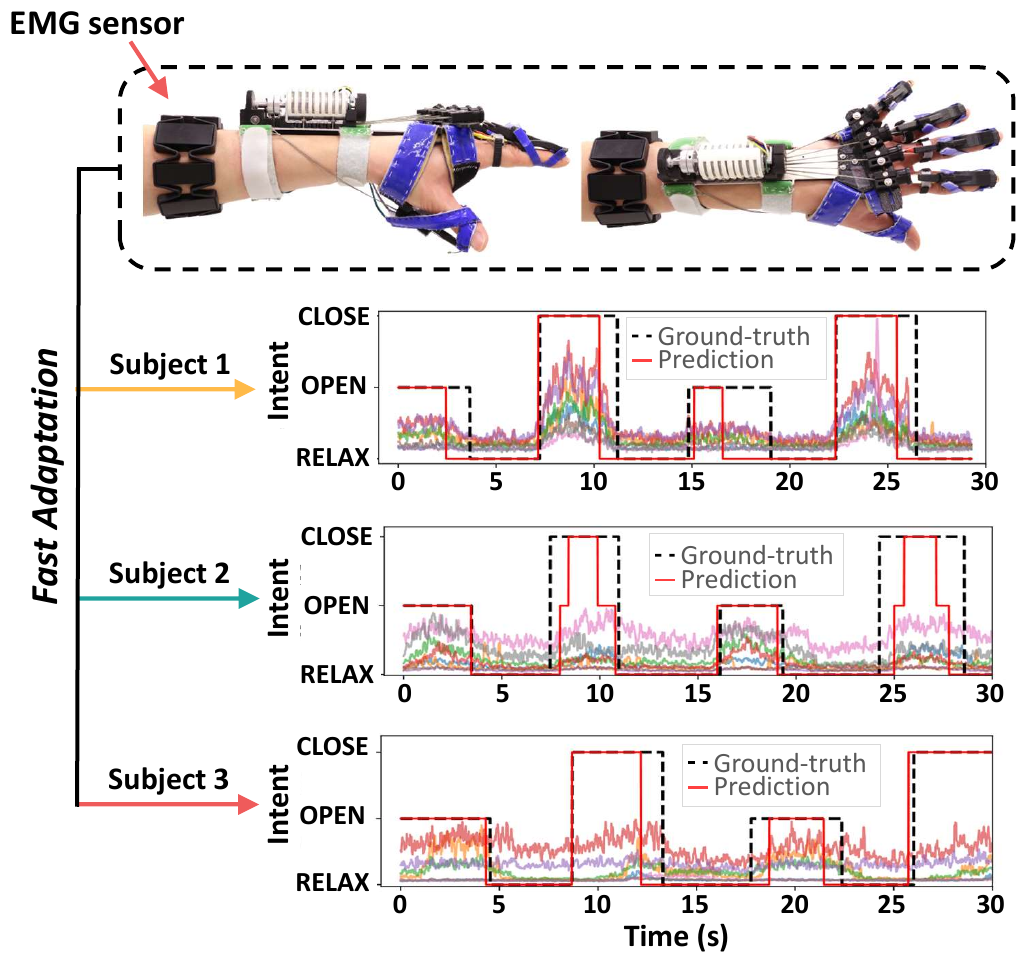}
    \caption{\textbf{MetaEMG for fast adaptation on three subjects with different EMG patterns.} Our method trains EMG-based intent inferral models that can quickly and efficiently adapt to new subjects in the context of a wearable robotic orthosis. Shown are EMG signal recordings of three different stroke survivors, and each colored series in the plots represents the reading of one of the eight EMG electrodes. Training models on new stroke subjects is difficult due to the large subject-to-subject variation in EMG signaling.}
    \label{fig:teaser}
\end{figure}

In this paper, we propose a meta-learning (or learning to learn) approach to training intent inferral models on a robotic orthosis. With this approach, a model can be explicitly optimized to adapt to new environments and tasks when labeled training examples are limited. In the context of wearable robotics, this means training intent inferral models that can quickly adapt to new subjects and new sessions with as little subject- or session-specific data as possible. Our directional goal is to train models that are as effective as possible in making use of the limited data available for any new patient or session. 
% \jx{shall we say new condition or session? We should be consistent. We did not run cross-condition experiments specifically.}

Specifically, we leverage real clinical data collected from different subjects and different orthosis use conditions in order to train a base model for intent inferral using Model-Agnostic Meta-Learning (MAML)~\cite{finn2017model}. The MAML algorithm explicitly rewards a base model that is easily adaptable when presented with new data. When the intent inferral model must be used for a new patient or on a new day, we only need a small set of new, session-specific training data to fine-tune the base model to the new condition. We find that our method, which we dub \OURS, is particularly effective at learning good intent classifiers with only a few training epochs, and that classifiers trained via \OURS are better at adapting to new subjects. In summary, our main contributions are as follows:

\begin{itemize}
    \item To the best of our knowledge,
    we are the first to formulate intent inferral on stroke subjects as a meta-learning problem. Our proposed method, \OURS, achieves fast adaptation to a new session or subject with only a few training samples and epochs. 
    \item We evaluate the performance of our method on data collected from five stroke subjects. We show that our method outperforms baselines whose models are not meta-learned or are only trained on the session or subject-specific data.
\end{itemize}
\section{RELATED WORK}

\subsection{EMG-based Intent Inferral}
Intent inferral is the process by which the device collects biosignals from the user, and uses these signals to infer the activity that the user intends to perform so that it can provide the right type of physical assistance. Machine learning has become an increasingly important approach in intent inferral with EMG signals. A majority of works~\cite{xu2022adaptive,meeker2017emg,batzianoulis2018decoding,sensinger2009adaptive,amsuss2013self,chen2013application,zhang2013adaptation,he2012adaptive} in this area rely on classical machine learning models, such as Linear Discriminant Analysis (LDA), Random Forests (RF), or Support Vector Machines (SVM). While having the advantages of being easily trackable and sometimes having closed-form solutions, these models lack the ability to model complex temporal relationships or non-linearity and often work better with hand-crafted feature engineering. 

On the other hand, more works~\cite{buon201} have started employing deep learning models such as convolutional neural networks (CNNs) for this application. Several works in the past decade have highlighted the benefit of neural networks over traditional machine learning approaches in EMG tasks, including gesture recognition~\cite{park2016} and inter-session adaptation~\cite{ameri2019deep}. Deep learning has also been shown to mitigate concept drift caused by the physical shifting of EMG sensor electrodes. Both Amari et al.~\cite{ameri2019deep} and Sun et al.~\cite{sun2022deep} showed that transfer learning with deep neural networks can be used to help mitigate concept drift from electrode migration. While high-capacity neural networks are better at modeling complex relationships between data and labels, they normally require more training data to realize their full power, which is extremely hard to collect on a large scale in our work with stroke subjects.

\subsection{Meta-learning with Biosignals} Meta-learning has been shown in biomedical research to be an effective framework to fast adapt high-capacity models to new individuals, reducing the need to collect large-scale subject-specific training data. Different from the group of works on semi-supervised learning~\cite{xu2022adaptive,zhai2017self,sensinger2009adaptive,amsuss2013self,chen2013application,edwards2016application,zhang2013adaptation,he2012adaptive,liu2015towards}, which requires user-defined heuristics for labeling unlabelled data, meta-learning pretrains the model on offline data in a way that allows it to quickly adapt to a new distribution. In MetaPhys~\cite{prorokovic2020meta}, the authors train a model to predict heart measurements from video data alone, and they use meta-learning to show how the models trained on a group of individuals can be fine-tuned on a smaller dataset to adapt to a new unseen individual. Similarly, both MetaSleepLearner~\cite{banluesombatkul2020metasleeplearner} and Prorokovich et al.~\cite{prorokovic2019} utilize meta-learning to adapt models to new individuals, in the contexts of sleep stage classification and speech recognition respectively. These works all focus on healthy subjects while we are working with stroke subjects, which is a much more challenging task due to their abnormal muscle activation and spasticity. 

Similar to ours, Prorokovich et al.~\cite{prorokovic2020meta} also work with disabled-bodied subjects (amputees), and they employed meta-learning for orthosis recalibration in the context of proportional control. They sought to train a regression model which predicts the strength of the movement generated by the user. In this work, the authors train regressors on data from both healthy subjects and amputees. However, in their work, the training and testing tasks contain the same individuals performing the same movements, which is different from our work, which seeks to adapt classifiers to new subjects and sessions. To the best of our knowledge, we are the first to use meta-learning to mitigate the burden of data collection needed to adapt high-capacity neural networks on stroke subjects.

\begin{figure}
    \centering
    \includegraphics[width=1\linewidth]{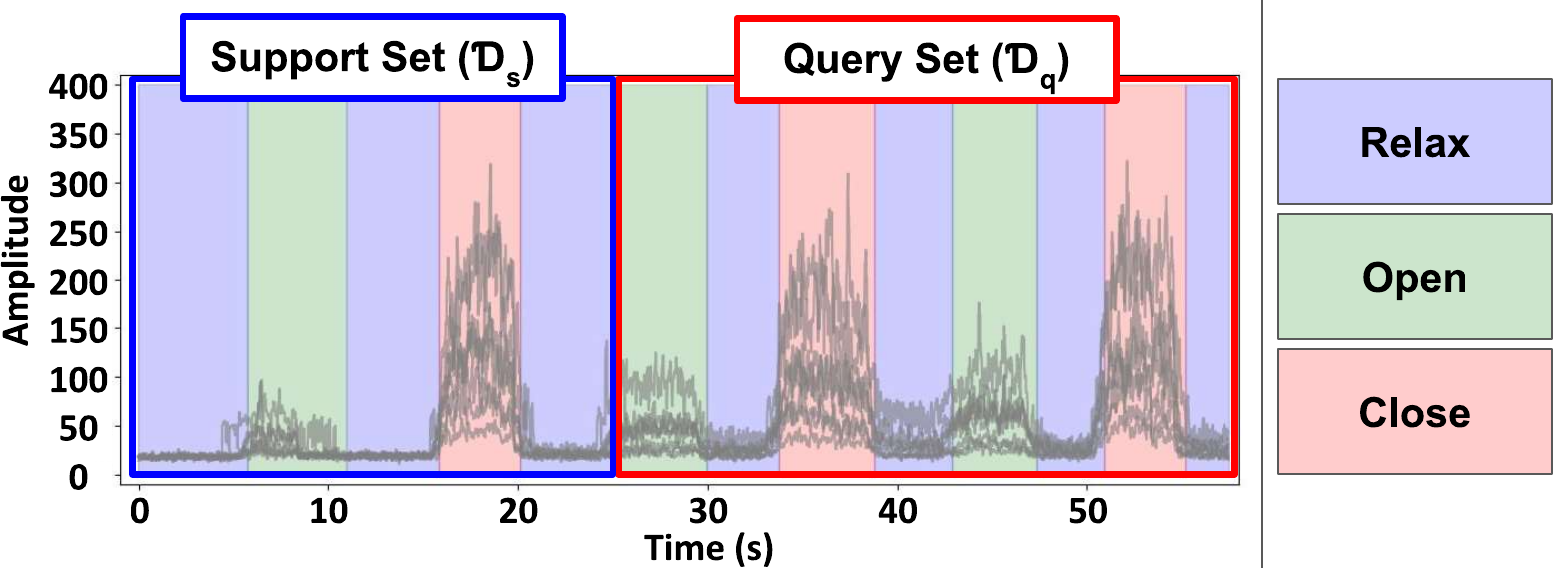}
    \caption{\textbf{Task visualization in EMG intent inferral}. We define an EMG task as a single uninterrupted recording with the 8-channel EMG armband. During each recording, users close and open their hands three times. The first open-relax-close motion (outlined in blue) is the support set. The third and second open-relax-close motions (outlined in red) consist of the query set. The ground-truth intent (verbal cues) is shaded in blue, green, and pink for relax, open, and close, respectively.}
    \label{fig:task}
\end{figure}

\begin{figure*}
    \centering
    \includegraphics[width=1\linewidth]{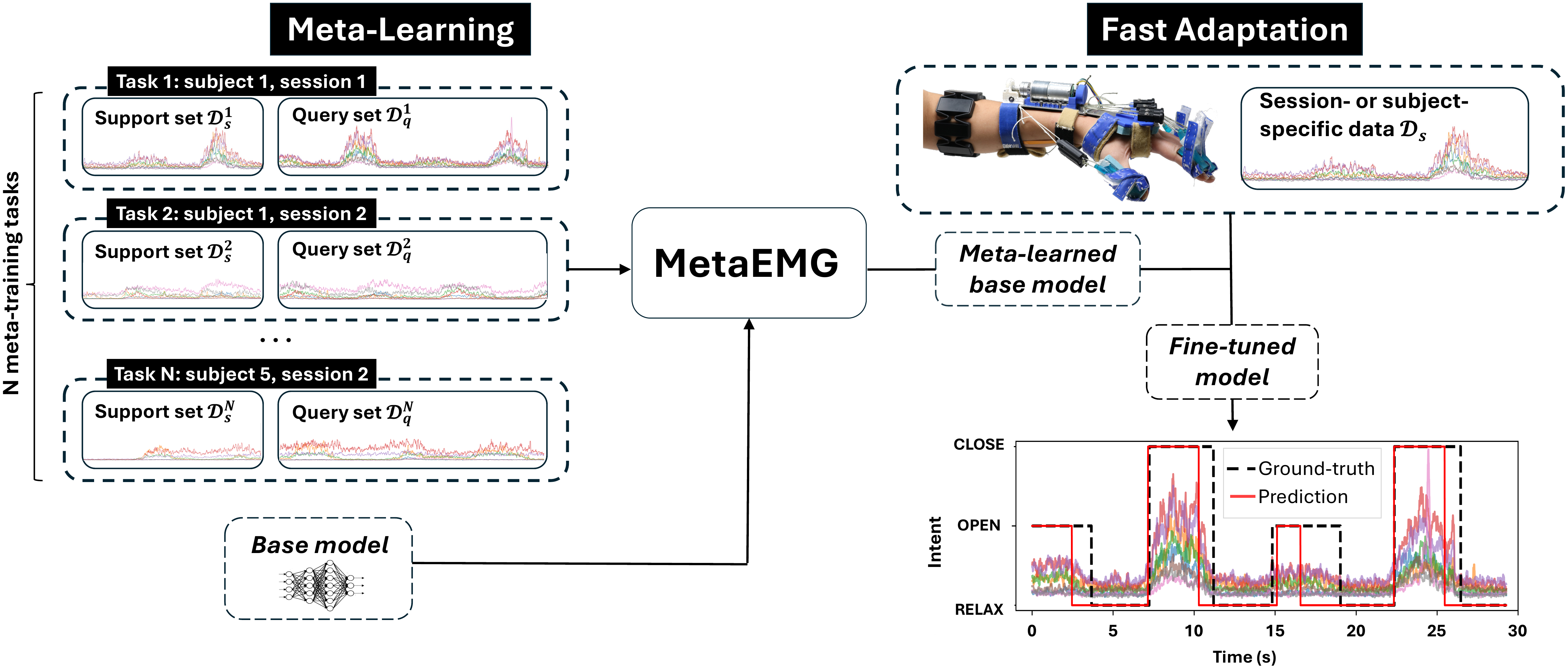}
    \caption{\textbf{\OURS for EMG intent inferral}. Our method uses previously collected data from our orthosis to meta-learn models that adapt to new sessions or subjects more quickly and with less data. }
    \label{fig:overview}
\end{figure*}
\section{Method}

% We believe that formulating stroke intent inferral as a meta-learning problem has the potential to mitigate the data collection burden when adapting to a new patient or session. At the core of our method, we design the intent inferral tasks in our meta-learning formulation to simulate the realistic use cases, and through meta-learning, the base model can reach a high classification accuracy with a very limited dataset and very few fine-tuning epochs at unseen tasks. 

\subsection{Intent Inferral Formulation}

We formalize intent inferral as the forecasting problem of predicting the intended user movement (in our case, hand opening, closing, or relaxing) from a stream of biosignals collected by a wearable robot (in our case, EMG data collected by an armband). Specifically, our goal is to determine intent from a data vector $x_t$ where at every time $t$, $x_t$ stores a 2-second window of EMG data $e^i$ across all 8 channels of our sensor:
$$x_t=[e^1_t, e^2_t...e^8_t]^T$$
Here, we specifically aim to predict which movement out of $\{open,close,relax\} $ the subject intends to perform with their hand, and we achieve this by calculating a vector of intent probabilities:
$$\hat{y_t}=[p^R_t, p^O_t, p^C_t]^T$$
where $p^R_t$, $p^O_t$, and $p^C_t$ represent the probabilities that the user’s intent is to relax (i.e. maintain current posture), open and close their hand, respectively.

We consider a base intent inferral model as a neural network parameterized by parameters $\theta$ whose output is used to derive the three intent probabilities, or $p^R_t$, $p^O_t$, and $p^C_t$. Traditionally, such a model would be trained on a dataset of labeled data collected by the experimenter at the beginning of each patient session. However, our goal is to reduce the burden of data collection and training for every session of use, so that we can more easily onboard new subjects and new therapeutic activities with our device. 

\subsection{Intent Inferral as a Meta-Learning Problem}

Meta-learning is characterized by the use of meta-training and meta-testing tasks to guide the learning process toward representations best suited for fast adaptation. Meta-training tasks refer to the tasks that are used in the meta-learning procedure to find a good base model initialization, and meta-testing tasks are unseen holdout tasks to evaluate the adaptation ability of the meta-learned base model. A task is defined by a tuple $\mathcal{T}=(\mathcal{D}_s,\mathcal{D}_q) $ composed of a support set $\mathcal{D}_s$ and a query set  $\mathcal{D}_q$, where $\mathcal{D}_s$ is much smaller in size than  $\mathcal{D}_q$ and contains only a small number of labeled data points. The objective of meta-learning in the context of supervised learning is to minimize the cross-entropy loss $\mathcal{L}$ of the meta-learned base model on the query sets of the meta-testing tasks.

In EMG intent inferral, we define a task as a continuous uninterrupted recording of EMG sensor data. For each recording, we instruct the subjects to open and close their hands three times by giving verbal cues of open, close, and relax. Each verbal cue lasts for 5 seconds and there is a relax cue between each open and close cue. The recording contains the EMG signals and their corresponding cues as the ground truth intent labels. We define each completion of the opening and closing hand as one \textit{open-relax-close motion}. 
We define the support set $\mathcal{D}_s$ to be the EMG signals of the first motion. In contrast, the query set $\mathcal{D}_q$ contains the second and third rounds of open-relax-close motions. In other words, each EMG task is given by $\mathcal{T}=(\mathcal{D}_s,\mathcal{D}_q)$ such that $\mathcal{D}_s=\{(x_t,y_t)\}_{t={1}}^{k}$ and $\mathcal{D}_q=\{(x_t,y_t)\}_{t={k+1}}^{n}$, where $k$ denotes the end of the first open-relax-close motion and $n$ denotes the end of the entire recording. An example of our intent inferral task is shown in Fig.~\ref{fig:task}.

With this task structure, the meta-objective can be interpreted as follows. We seek to train a base model using the complete support set $\mathcal{D}_s$ and query set $\mathcal{D}_q$ over a number of meta-training tasks. Then, when presented with a new meta-testing task (e.g., an unseen task), this model should be quickly fine-tuned using only the support set $\mathcal{D}_s$ of the new task, meaning just a single round of open-relax-close motion. 

\subsection{MetaEMG}

Under our task formulation, the proposed MetaEMG is an application of the Model-Agnostic Meta-Learning (MAML)~\cite{finn2017model} algorithm to EMG intent inferral. MAML is a meta learning algorithm that focuses on developing models with enhanced adaptability. Rather than optimizing a model for a specific task, MAML aims to create a model that can quickly adapt to new tasks with minimal additional training essentially training the model to be an efficient learner. By doing so, this algorithm can produce models that perform well across a diverse range of tasks, even when provided with only a small number of examples for each new task~\cite{finn2017model}.

MetaEMG aims to find the optimal initial model parameters for adapting a base model to a new session or subject from a limited dataset in only a few fine-tuning epochs. In MetaEMG, two nested optimization loops are used to fine-tune a base model for different tasks using small amounts of data, as shown in Fig.~\ref{fig:overview} and Alg.~\ref{alg:algo1}.

\begin{algorithm}
\caption{MAML for EMG intent inferral}\label{alg:algo1}
\begin{algorithmic}[1]
\Require $T$: collection of $N$ tasks $\tau_i$, where each $\tau_i$ is a single session recording
\Require $\alpha, \beta$: learning rates for inner and outer optimization loops
\Require $M$, $K$: epochs for inner and outer optimization loops
\State randomly initialize base model parameters $\theta$
\State $k,m \gets 0$
\While{ $k \leq K$ } \textit{(outer loop)}
    \For{each task $\tau_i$}
    \State $\hat{\theta_i} \gets \theta$
    \State extract support and query sets $\mathcal{D}_s^i$ and $\mathcal{D}_q^i$
    \While{ $m \leq M$} \textit{(inner loop)}
        \State $\hat{\theta_i}$: $\hat{\theta_i} \gets \hat{\theta_i} - \alpha\nabla_{\hat{\theta_i}}\mathcal{L}_{\mathcal{D}_s^i}(f_{\hat{\theta_i}})$
    \EndWhile
    \State evaluate query loss $\mathcal{L}_{\mathcal{D}_q^i}(f_{\hat{\theta_i}})$
    \State evaluate and store gradients $\nabla_{\theta}\mathcal{L}_{\mathcal{D}_q^i}(f_{\hat{\theta_i}})$
    \EndFor
    \State update $\theta$: $\theta \gets \theta - \beta\nabla_{\theta}\sum_{i=0}^N \mathcal{L}_{\mathcal{D}_q^i}(f_{\hat{\theta_i}}) $
\EndWhile
\end{algorithmic}
\end{algorithm}

One of the advantages of formulating intent inferral in the meta learning framework is that it allows us to train models in a way that directly mirrors their use case. In the inner loop of MetaEMG, a base model is fine-tuned on the support set $\mathcal{D}_s$ of a training task,
and that base model's ability to adapt to the task is measured through the query loss $\mathcal{L}_{\mathcal{D}_q}$ of that task (Alg.~\ref{alg:algo1}, steps 7 - 10). The query loss is simply the cross-entropy loss for the query set of the task computed between the predictions and ground-truth labels. This process simulates the real-life use case of fine-tuning an intent classifier on a small amount of demonstration data and using that model for real-time inference in a session with the device. 

Moreover, the base model update is performed once per iteration of the outer optimization loop. The query loss is used to accumulate the gradients on the base model parameters across all of the training tasks until a backward step is performed once all of those training tasks have been seen (Alg.~\ref{alg:algo1}, steps 11 - 13). 

Importantly, the fine-tuning in the inner loop is performed via gradient descent on $\mathcal{L}_{\mathcal{D}_s}$ (taken with respect to the latest parameter set $\hat{\theta_i}$), while the base model update of the outer loop is performed via gradient descent on $\mathcal{L}_{\mathcal{D}_q}$ (taken with respect to the base model parameter set $\theta$) which requires backpropagation through the inner optimization and thus higher-order gradients. By updating the base model using $\mathcal{L}_{\mathcal{D}_q}$, the training process favors base model updates that allow it to on average generalize to both the support $\mathcal{D}_s$ and query $\mathcal{D}_s$ sets of the training tasks despite only being fine-tuned on the support sets. For EMG intent inferral, we hypothesize that this training structure can produce intent classifiers that are less likely to overfit when fine-tuned to small amounts of demonstration data.

Models are trained on a collection of meta-training tasks and then evaluated on a collection of meta-testing tasks, and the specific allotment of session recordings in each collection depends on the individual use case for the model that we describe in the experimental setup. Generally, larger diverse meta-training task collections are favored, and in this paper, we select recordings to simulate realistic use cases to evaluate our meta-learning framework's fast adaptation ability.

\section{Data Collection and Session Conditions}
\label{sec:data}

%We evaluated our proposed method by training models using MAML and assessing how well they are able adapt to new tasks. We defined two specific use-cases for our method, subject adaptation and session adaptation, and we generated different train/test task collections to experimentally evaluate our models on these use-cases. Importantly, all of our data comes from real stroke survivors and was collected using the Myhand orthosis, a robotic assistive device previously developed in the lab.

All data used in this study was collected using MyHand, a robotic exotendon device comprised of a system of motors and linkages which assist the user in opening and closing their hand, as shown in Fig.~\ref{fig:teaser}. The device rests on the user's forearm and through a series of tendons connected to the user's fingertips assists the hand in opening and closing a grasp. EMG sensor data is recorded from an 8-electrode armband (Myo, Thalmic Labs). Once a command is issued to either open or close the hand (for example, by an intent inferral model as the one that is the focus of this study), the orthosis engages the motor and either retracts or extends the exotendons on the dorsal side of the finger. Finger extension is actively assisted by the device through tendon retraction, while finger flexion is allowed unimpeded via the subject's own strength. Additional hardware details can be found in our previous studies~\cite{park2018multimodal,chen2022thumb}. 

% \begin{figure}
%     \centering
%     \includegraphics[width=0.9\linewidth]{figures/hardware_fig.pdf}
%     \caption{Visualization of the Myhand orthosis and EMG armband. MyHand assists in grasping by providing mechanical support in opening and closing the user's hand.}
%     \label{fig:hardware}
% \end{figure}
 
\subsection{Data}

Critically, the fact that we collect our data using a complete, functional assistive device means that our collection protocol can span the conditions that are meaningful for real-life operations. Our previous work~\cite{meeker2017emg,xu2022adaptive} has shown that, in our patient population, EMG signals display significant differences when collected in isolation (i.e. using just an armband and no other hardware) vs. when collected in conjunction with an active orthosis assisting finger movement. Thus, in our method, we collect data in four different experimental conditions as follows:
\begin{enumerate}
    \item \textbf{Arm on table, motor off}: the assisted arm of the subject rests on the table, and the motors of the MyHand are disengaged, providing no physical assistance to the subject.
    \item \textbf{Arm on table, motor on}: the assisted arm of the subject rests on the table, and the motors of the MyHand are engaged, allowing the device to provide active grasp assistance to the user.
    \item \textbf{Arm off table, motor off}: the subject keeps their arm raised to shoulder level for the duration of the session, and the motors are disengaged, providing no physical assistance to the subject.
    \item \textbf{Arm off table, motor on}: the subject keeps their arm raised to shoulder level for the duration of the session, and the motors of the MyHand are engaged, allowing the device to provide active grasp assistance to the user.
\end{enumerate}

\subsection{Pre-processing}

The raw output of the EMG sensor is an 8-channel time-series signal (shown in Fig.~\ref{fig:teaser}) sampled at 100 Hz which is pre-processed before going into our intent inferral model. The labels attached to each sample are the verbal cues given to the subject during the session, and in a 2-second time interval, there are 200, 8-channel samples. During pre-processing, the 8-channel EMG signal is clipped to the range $[0,1000]$ to discard any outliers, and each channel is re-scaled to the range $[-1,1]$. Finally, the scaled 8-channel signal is binned into 2 second-long 8-channel data windows, following a sliding window stride length of 10ms. While muscle activation is known to occur in a much smaller time window, we hypothesize that 2 seconds encapsulates the total reaction time from interpreting the verbal command to actual movement exertion.

\section{Experimental Setup}

\begin{table}[t]
\centering
\caption{\textbf{Subject information.}}
 \begin{tabular}{c c c c} 
 \toprule
 Subject ID & Age & Gender & FM-UE \\
 \midrule
 S1 & 46 & Male & 27 \\ 
 S2 & 48 & Male & 26 \\
 S3 & 53 & Female & 26 \\
 S4 & 31 & Male & 50 \\
 S5 & 53 & Male & 47 \\ 
 \bottomrule
 \end{tabular}
 \label{tab:subjects}
\end{table}

We evaluate \OURS in a series of experiments conducted on data collected using the MyHand orthosis and involving five chronic stroke survivors of moderate muscle tone (MAS $\le 2$) and varying degrees of arm impairment (FM-UE scores listed in Table~\ref{tab:subjects}) in sessions approved by the Columbia University Institutional Review Board (IRB-AAAS8104). Subjects are generally able to actively close their hands but not open them, which is one of the motivating factors in both the robotic orthosis design as well as the intent inferral structure.

In each single session, the subjects are asked to follow verbal cues asking them to either relax, close, or open their hand. Each cue lasts a duration of five seconds, and this procedure is followed across the four experimental conditions outlined in Sec.~\ref{sec:data}. Each recording contains three rounds of open-relax-close motions. In total, 14 recordings were gathered for each subject over the course of two separate days, where 8 of those recordings came from the first day of data collection and 6 came from the second day. In order to study intersession concept drift, the two days are at least a week apart, providing enough time for the drift caused by the chronic change in muscle condition to happen. For consistency with our method's terminology, we refer to individual recordings as tasks such that the first open-relax-close motion of that recording is encapsulated in its support set $\mathcal{D}_s$ and the remainder of the recording in the query set $\mathcal{D}_q$.

\subsection{Assessment Scenarios}

All of the experiments are conducted using this same dataset of five subjects. In each experiment, models are first pretrained on the meta-learning task collection and then evaluated on the meta-testing tasks. For both meta-learning and meta-testing tasks, $\mathcal{D}_s$ is always used to fine-tune a base model to a task. The distinction is that during the pretraining phase, the fine-tuned model's prediction error on $\mathcal{D}_q$ is used to guide the optimization of the base model, whereas in the evaluation phase, the classification accuracy on the task's $\mathcal{D}_q$ is used as our primary metric of model performance. We evaluate the performance of our method under two assessment scenarios as follows. 

\begin{itemize}
    \item \textbf{Session adaptation.} This scenario pertains to intersession concept drift, and it seeks to simulate using the orthosis on a subject seen previously in a different session on a different day. A session is defined as a single-use between donning and doffing the device. We run a single experiment, and our meta-learning and meta-training tasks are simply grouped by day. Specifically, the meta-learning task collection contains recordings of all five subjects across all 4 data collection conditions, but it only contains those recordings collected on the first day of data collection. The meta-testing task collection is comprised of the recordings from only the second day of data collection, but it also includes all five subjects and all data collection conditions.
    \item \textbf{Subject adaptation.} This scenario seeks to address the challenge of onboarding new subjects onto our robotic orthosis. To simulate this scenario, we conduct five separate experiments, each one simulating the onboarding of one held-out subject given that we have seen the other four. In each experiment, all 14 recordings of the held-out subject are used as the meta-testing tasks, and the meta-training tasks are comprised of all of the session recordings from the other four subjects.
\end{itemize}

% \mc{This is where you very clearly and concretely explain how we test: we train on this, test on that, the metric is such-and-such, and so on. It needs to be tied back into, and use the language introduced in, allprevious sections, namely the support and query sets from Sec. III, the four conditions listed in Sec. IV, etc.}

\subsection{Model Architecture}

% \mc{Here is where you can add details about layers, neurons, softmax, etc.} 

The base architecture for our models is a simple 3-layer fully connected neural network of 512, 128, and 3 neurons at each layer, respectively. During inference, the final layer output of our network is passed through a softmax activation function to obtain a vector of intent probabilities. Our models and datasets are developed in PyTorch and trained on an NVIDIA GeForce RTX 3090 GPU for 50 epochs on a learning rate scheduler, reducing the outer learning rate by 0.9 every 10 epochs. We use Adam as the optimizer for both optimization loops, and we find the combination of 0.0005 outer learning rate and 0.0001 inner learning rate to yield the most stable training.

\subsection{Baselines}
In order to evaluate \OURS as an adaptation method, we compare it to several baselines without pretraining or with vanilla pretraining. The fast adaptability of \OURS allows it to be used as a base model for continuous and lifelong online learning with higher-capacity models. We are mostly interested in the performance comparison under limited training epochs as our main goal is to achieve fast adaptation to a new subject or session. However, we also include baselines that are trained to convergence, which normally takes much longer training time. These converged baselines are not suitable to be used as base models for online updates but we still include them for the completeness of results. 

All of our baselines have the same base intent inferral model neural network architecture, and the evaluation metric used across the board is the average classification accuracy on the $\mathcal{D}_q$ of the meta-testing tasks. The methods that we evaluate are as follows.

\subsubsection{No-pretraining (3 epochs)} This baseline does not pretrain on any of the offline corpus of data from the meta-training tasks. We initialize the weights of the model randomly, train it directly on the support set $\mathcal{D}_s$ of the meta-testing tasks for 3 epochs, and evaluate its classification accuracy on the query set $\mathcal{D}_q$ of those same tasks. This helps us distinguish whether the pretraining process (meta-learning or vanilla pretraining) is indeed helping our models learn more effectively or if the training on the limited one open-close-relax motion from the new subject or session suffices to produce a good classifier.

\subsubsection{No-pretraining (converged)} This is the same as the previous baseline, except it is trained to convergence during the fine-tuning process on the support sets of meta-testing tasks, which takes 50 epochs. 

% Our first two baselines consist of no pre-training, and they generally help us distinguish whether the meta-learning process is indeed helping our models learn from a demonstration dataset more effectively or if the fine-tuning process in itself is enough to adapt to new tasks. No pre-training means that we take a model with randomly initialized weights, train it on the support set $\mathcal{D}_s$ of the training task, and evaluate its classification accuracy on the query set $\mathcal{D}_q$ of that same task. The \textit{No pre-training baseline} is trained on the meta-testing task's $\mathcal{D}_s$ for only 3 epochs, and the \textit{No pre-training baseline*} is trained on this same $\mathcal{D}_s$ to convergence.

\subsubsection{Conventional-pretraining (3 epochs)} This baseline aggregates both the support sets and the query sets of all meta-training tasks into a single dataset and treats them all as a single support set. The model is trained on the entirety of this lumped dataset, and it is fine-tuned to the support sets of the meta-testing tasks. 

\subsubsection{Conventional-pretraining (converged)} This is the same as the previous baseline, but it is fine-tuned to convergence on the support sets of the meta-testing tasks for 3 epochs. 

\subsubsection{MetaEMG} This is our proposed method, only fine-tuned for 3 epochs on the meta-testing tasks. We note that 3 epochs is convergence for models trained via MetaEMG.

% The second batch of baselines are defined by the conventional pre-training approach of aggregating all of the meta-training tasks into a single task and training a model on the entirety of this lumped task. This trained model is then fine-tuned on the $\mathcal{D}_s$ of the meta-testing task, and it is then evaluated on its classification accuracy on the meta-testing task's $\mathcal{D}_q$. Similar to above, \textit{Conventional pre-training} is fine-tuned on $\mathcal{D}_s$ for only 3 epochs, and \textit{Conventional pre-training*} is fine-tuned on $\mathcal{D}_s$ to convergence.

\section{Results \& Discussion}
\begin{table*}[ht]
\centering
    \caption{\textbf{Session adaptation results.} Classification accuracy and standard deviation across 5 stroke subjects. Results are averaged across 3 separate randomly generated training seeds. The best results for each subject are reported in bold.}
    \begin{tabular}[t]{c|ccccc|c}
        \toprule
         \textbf{Method} & \textbf{S1} & \textbf{S2} & \textbf{S3} & \textbf{S4} & \textbf{S5} & \textbf{Average}\\
        \midrule
        No-pretraining (3 epochs) & 52.9 $\pm$ 1.2\% & 41.1 $\pm$ 5.5\% & 53.1 $\pm$ 8.5\% & 55.8 $\pm$ 0.9\% & 57.1 $\pm$ 2.5\% & 52.06\%\\
        No-pretraining (converged) & 67.5 $\pm$ 0.6\% & 62.2 $\pm$ 0.7\%	& 73.7 $\pm$ 1.7\% & \textbf{79.6 $\pm$ 0.2\%} & 66.0 $\pm$ 2.2\% & 69.85\%\\
        Conventional-pretraining (3 epochs) & 70.2 $\pm$ 2.8\% & 63.3 $\pm$ 2.6\% & 76.3 $\pm$ 2.9\% & 71.6 $\pm$ 1.3\% & 71.3 $\pm$ 2.4\% & 70.59\%\\
        Conventional-pretraining (converged) & 70.3 $\pm$ 2.0\% & \textbf{64.7 $\pm$ 1.6\%} & 79.5 $\pm$ 2.1\% & 75.8 $\pm$ 0.8\% & 70.4 $\pm$ 2.9\% & 72.19\%\\
        \textbf{\OURS} & \textbf{74.5 $\pm$ 2.5\%} & 63.6 $\pm$ 0.9\% & \textbf{80.5 $\pm$ 1.0\%} & 78.6 $\pm$ 1.3\% & \textbf{74.4 $\pm$ 0.3\%} & \textbf{74.37\%} \\
        \bottomrule
    \end{tabular}
    \label{tab:session}
\end{table*}

\begin{figure}
    \centering
    \includegraphics[width=1\linewidth]{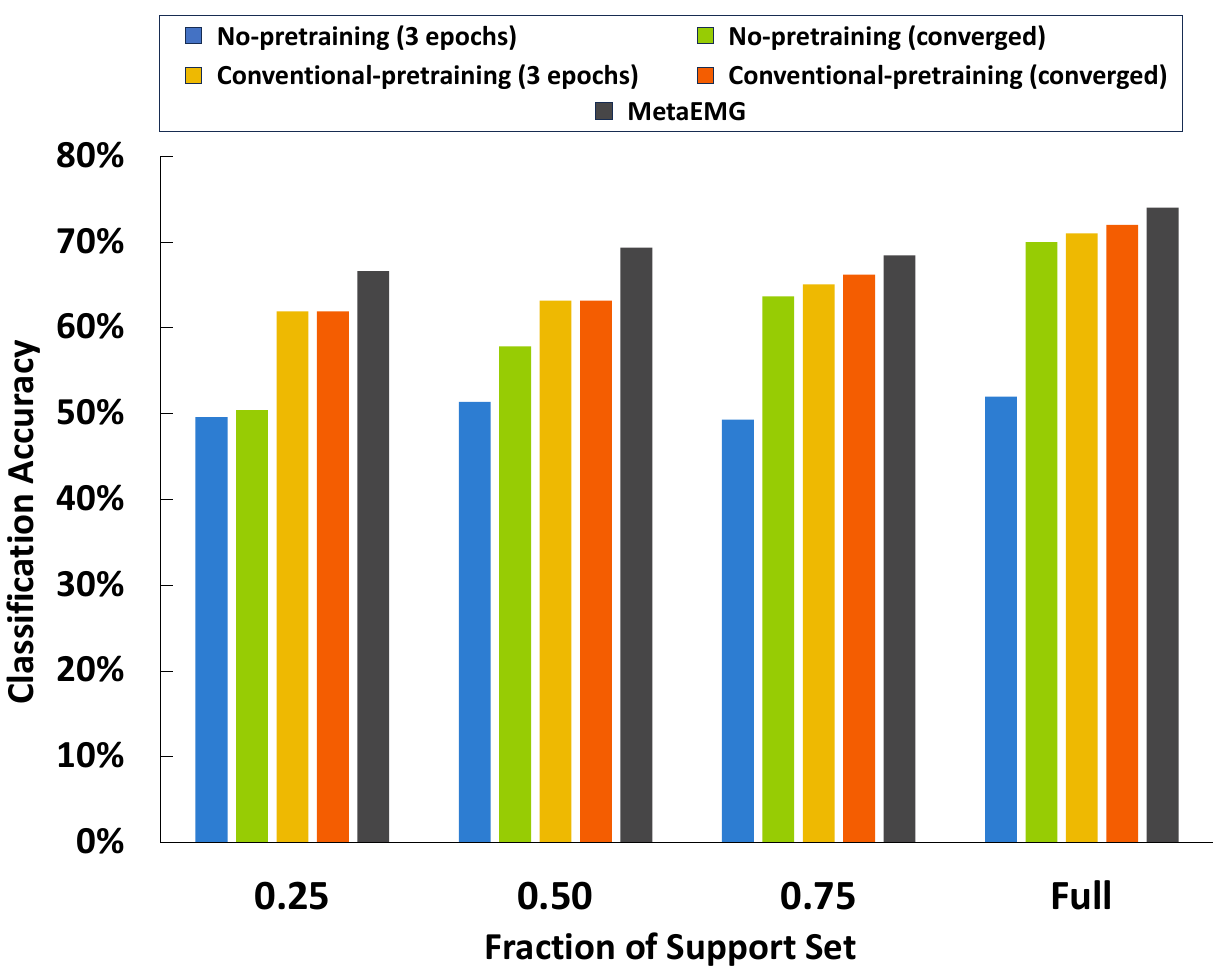}
    \caption{\textbf{Classification accuracy with fewer session-specific data.} In the session adaptation experiments, we further reduce the amount of session-specific fine-tuning data to 0.75, 0.5, and 0.25 of the original amount. }
    \label{fig:fewer_data}
\end{figure}

\begin{table*}[t]
\centering
    \caption{\textbf{Subject adaptation results.} Classification accuracy and standard deviation across 5 stroke subjects. Results are averaged across 3 separate randomly generated training seeds. The best results for each subject are reported in bold.}
    \begin{tabular}[t]{c|ccccc|c}
        \toprule
         \textbf{Method} & \textbf{S1} & \textbf{S2} & \textbf{S3} & \textbf{S4} & \textbf{S5} & \textbf{Average}\\
        \midrule
        No-pretraining (3 epochs) & 54.6 $\pm$ 1.5\% & 55.3 $\pm$ 1.6\% & 51.6 $\pm$ 5.1\% & 54.8 $\pm$ 2.9\% & 53.9 $\pm$ 1.0\% & 54.1\%\\
        No-pretraining (converged) & 60.1 $\pm$ 0.3\% & 66.8 $\pm$ 1.3\% & 68.3 $\pm$ 0.7\% & 74.7 $\pm$ 0.5\% & 69.4 $\pm$ 0.4\% & 67.9\%\\
        Conventional-pretraining (3 epochs) & 60.9 $\pm$ 0.9\% & 68.8 $\pm$ 1.0\% & 73.7 $\pm$ 0.7\% & 75.4 $\pm$ 1.7\% & 64.7 $\pm$ 1.4\% & 68.7\%\\
        Conventional-pretraining (converged) & 62.9 $\pm$ 0.7\% & \textbf{71.3 $\pm$ 0.4\%} & \textbf{74.2 $\pm$ 0.1\%} & 78.3 $\pm$ 0.6\% & 66.6 $\pm$ 1.3\% & 70.6\%\\
       \textbf{\OURS} & \textbf{63.8 $\pm$ 0.8\%} & 69.0 $\pm$ 1.1\% & 74.1 $\pm$ 0.5\% & \textbf{80.5 $\pm$ 1.1\%} & \textbf{70.0 $\pm$ 0.5\%} & \textbf{71.5\%}\\
        \bottomrule
    \end{tabular}
    \label{tab:subject}
\end{table*}%

\begin{figure}
    \centering
    \includegraphics[width=1\linewidth]{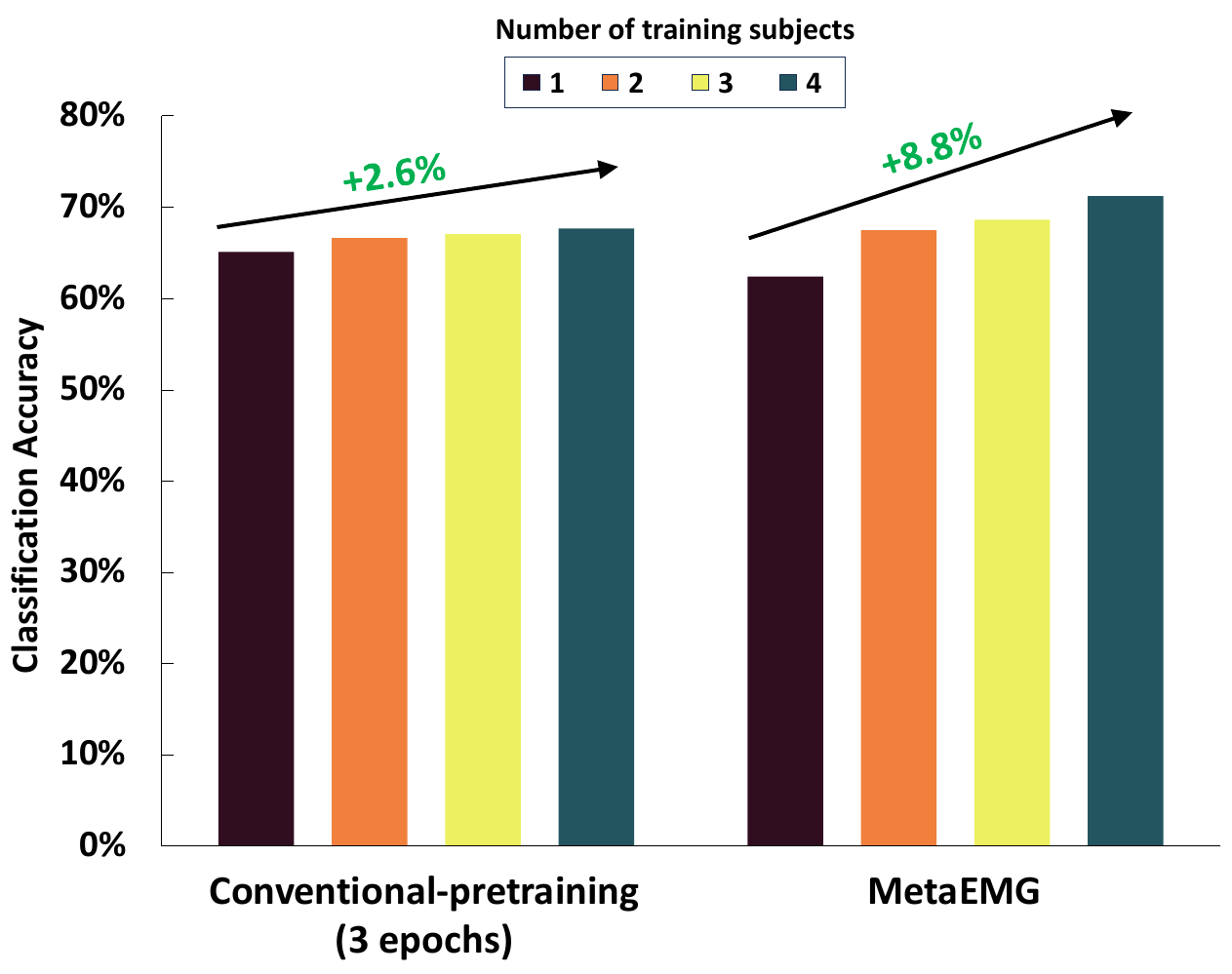}
    \caption{\textbf{Classification accuracy with different number of pretraining subjects}. Models are pretrained on 1, 2, 3, and 4 subjects before being fine-tuned on a subject not seen in pretraining. For each number of pretraining subjects, we run experiments with all possible partitions of the meta-training and meta-testing subjects and report the average classification accuracy.}
    \label{fig:fewer_subjects}
\end{figure}    

Through the experimental results from two realistic adaptation scenarios, we aim to demonstrate the fast adaptation ability of \OURS from two perspectives: (1) \OURS adapts better with very limited subject- or session-specific data, and (2) \OURS adapts better with only a few fine-tuning epochs.

Both perspectives are critical in intent inferral with EMG signals for stroke. Fast adaptation with less data relieves the burden of data collection for a new subject or session. Fast adaptation with fewer training epochs reduces the overhead time for using the device, and more importantly, it has the potential to be used as a base model for on-device online learning, which requires fast continuous model updates.

\subsection{Session Adaptation}

% In order to evaluate whether \OURS can achieve useful subject-to-subject knowledge transfer, we set up an experiment using data from the five subjects outlined in table II to determine if onboarding new subjects is any easier with \OURS than with conventional pre-training. In our subject-to-subject knowledge transfer experiment, we assign 1 of the 5 subjects in our database as the "new" subject and the rest as "known" so that the session recordings from the "known" and "new" subjects are collated into the meta-training and meta-testing task collections respectively. In this scenario, pre-training is performed on the 4 "known" subjects, and the pre-trained model is evaluated on the "new" patient by fine-tuning on its support set (calibration data) for 3 gradient steps. This routine is repeated on all 5 subjects such that all 5 subjects are sampled as "new" subjects once, and the results for the five subjects across the 3 pre-training conditions are compiled into Table 2. \pl{I wonder if a table defining the meta training and meta testing tasks for cross-subject would be helpful here for breaking up this long paragraph...}

\OURS outperforms all baselines on the average intent inferral accuracy in the session adaptation experiments, as shown in Table~\ref{tab:session}. We simulate the scenario of the same subject returning for another use session of our device after at least a week. Only one open-relax-close motion of EMG signals is provided as the support set for the meta-testing tasks. 

Without pretraining on any of the offline corpus of data from previous subjects or sessions, \textit{No-pretraining} baselines perform worse than \textit{Conventional-pretraining} baselines. This shows that our high-capacity neural network classifier benefits from pretraining on a larger corpus of offline EMG data. However, \textit{Conventional-pretraining} baselines still perform worse than \OURS. This is due to the intersession concept drift after donning and doffing the device, which is caused by chronic hand function changes and device position discrepancies on the forearm. Through meta-learning, \OURS adapts more efficiently to these drifts in EMG signals, achieving a higher classification accuracy (74.37\%) compared to \textit{Conventional-pretraining (3 epochs)} (70.59\%).

We further investigate if \OURS adapts better with even smaller fine-tuning datasets. We downsample the support set of each meta-testing task to 25\%, 50\%, and 75\%, and we compare the average classification accuracies across all subjects (shown in Fig.~\ref{fig:fewer_data}). We find that \OURS is most resilient to the support set reduction, maintaining a classification accuracy above $66\%$, while other baselines' performance deteriorates more dramatically. This suggests that meta-learned models are potentially also better at learning from smaller session-specific datasets.

Apart from adapting better with limited session-specific data, another major advantage of \OURS is that it can adapt effectively with very few epochs. \OURS shows a large improvement in classification accuracy over the baselines when limited to only 3 epochs of fine-tuning for the meta-testing tasks. While at convergence, the difference between the baselines and \OURS is small. We note that on our hardware, it takes 50 epochs for both baselines to converge when fine-tuning on the session-specific data of the meta-testing tasks. The training time for convergence of the baselines is a magnitude longer than training \OURS for 3 epochs, yet \OURS still outperforms the converged baselines. Being able to adapt fast with a few epochs is beneficial for two reasons. (1) It is essential for continuous online updates, and the meta-learned base model has the potential to enable on-device life-long learning. (2) It enables the use of higher capacity models where each training epoch can take a significant amount of time. 

% the difference in adaptation time between the conventional pre-trained model fine-tuned to convergence and \OURS fine-tuned to only 3 epochs was around 10-fold. With our current dataset and models, this only constitutes a 10-second difference in training time, but this 10-fold difference would be much more consequential with a larger model or with additional sensor data inputs. \jx{rewrite}.

\subsection{Subject Adaptation}

Table~\ref{tab:subject} shows the subject adaptation results and \OURS is on average better than all of the baselines, suggesting that meta-learning yields benefits in subject adaptation over the other methods. Subject adaptation is generally a harder problem compared to session adaptation because the distributional shift across subjects is larger than that across different sessions from the same subject. This is shown by the performance drop of all methods from the session adaptation experiments except for \textit{No-pretraining (3 epochs)}. However, meta-learning is still able to produce a base model that adapts better with limited subject-specific data and very few epochs by transferring knowledge across similar subjects. 

% Our expectation was that pre-training our intent inferral models through meta learning would improve how fast they are able to learn from the new subject’s demonstration data since there is likely to be some amount of similarity between new subjects and subjects seen in the past. Our results show the  Importantly, with only 4 subjects present in the meta-learning task collection, the \OURS models outperformed both baselines at the 3-epoch adaptation length by a considerable amount. This suggests that meta-learning is indeed able to produce fast-adapting models. \jx{we need more high-level blue sky discussion here on why being able to fine-tune with 3 epochs is important.}

% In addition to assessing whether meta-learned models can adapt faster, we also sought to determine if meta learning could also produce more accurate models than our baselines could when fine-tuned to convergence. While fast adaptation alone is valuable to us because of the context of limited clinical time, we also want to  know if meta-models learn more powerful embeddings than they would via conventional pre-training methods. As seen in Table I, \OURS models do on average slightly outperform the baselines fine-tuned to convergence. While this improvement is small, we hypothesize that with more meta-training tasks of higher diversity the benefit of meta-learning will be greater.

In order to further understand if different pretraining methods would improve as we add more subjects and data diversity to our database, we conduct another ablation experiment where we vary the number of pretraining subjects presented in the meta-training task collection, as shown in Fig.~\ref{fig:fewer_subjects}. We compare \textit{Conventional-pretraining (3 epochs)} and \OURS with only 1, 2, 3, and 4 subjects in our meta-training tasks and with the rest of the subjects in the meta-testing tasks. \OURS shows a more significant improvement in intent inferral accuracy when the number of pretraining subjects increases, and we believe this indicates that with more subjects added to our corpus of offline data, the advantage of \OURS will become even more pronounced. 

% We note that these experiments are conducted using the same dataset of 5 stroke subjects, but in each condition we generate different meta-training and meta-testing task collections. The four conditions indicate 1, 2, 3, and 4 subjects present in the meta-training collection, and in every condition the meta-testing collection contains data from the remaining subjects. Figure 5 shows the results of this experiment, and it suggests a stronger trend for \OURS in classification accuracy vs number of training subjects. 

\section{Conclusions \& Future Work}

In this paper, we explore the use of meta-learning for pretraining intent classification models specifically for fast adaptation. We highlight two use cases for this approach using real stroke survivor data from a robotic orthosis, and we provide a detailed analysis of the method. Our experiments show that there is a strong benefit to using meta-learning as a pretraining method in intent classification, and we believe this approach extends to other domains of biosignals. More importantly, the fast adaptability of \OURS can enable continuous on-device life-long learning with higher capacity models in the future. To our knowledge, this is the first time meta-learning has been explored as a way to improve intent inferral in stroke survivors, and we hope that it inspires other health domains where large-scale data collection is difficult.

%                                   % the textheight of the last page by a suitable amount.
%                                   % This command does not take effect until the next page
%                                   % so it should come on the page before the last. Make
%                                   % sure that you do not shorten the textheight too much.

\section*{Acknowledgments}
This work was supported in part by the National Institutes of Health (R01NS115652, F31HD111301) and a National GEM Consortium Fellowship.

\bibliographystyle{IEEEtran} % We choose the "plain" reference style
\bibliography{references} % Entries are in the refs.bib file

\end{document}